\documentclass[letterpaper,conference]{IEEEtran}
\IEEEoverridecommandlockouts
\PassOptionsToPackage{bookmarks=false,hidelinks}{hyperref}
\usepackage{hyperref}
\usepackage[left=1.62cm,right=1.62cm,top=1.9cm]{geometry}
\usepackage{cite}
\usepackage{amsmath,amssymb,amsfonts}
\usepackage{algorithmic}
\usepackage{graphicx}
\usepackage{textcomp}
\usepackage{xcolor}
\usepackage{orcidlink}
\usepackage[linesnumbered, vlined, ruled]{algorithm2e}
\SetAlFnt{\footnotesize}
\SetAlCapFnt{\footnotesize}

\SetCommentSty{mycommfont}
\usepackage{wrapfig}
\usepackage{booktabs}
\usepackage{multirow}
\usepackage{array}

\usepackage{subcaption}
\begin{document}

\title{
Analysis of Deep Learning-Based Colorization and Super-Resolution Techniques for Lidar Imagery\\
}


\author{
    \IEEEauthorblockN{
        \vspace{1em}
        Sier Ha\,\orcidlink{0009-0000-3617-107X},
        Honghao Du\,\orcidlink{0009-0008-7600-0302},
        Xianjia Yu\,\orcidlink{0000-0002-9042-3730}, 
        Jian Song\,\orcidlink{0009-0009-4075-12169}, 
        Tomi Westerlund\,\orcidlink{0000-0002-1793-2694}
    }
    \IEEEauthorblockA{
        \normalsize
        \IEEEauthorrefmark{2}\href{https://tiers.utu.fi}{Turku Intelligent Embedded and Robotic Systems (TIERS) Lab, University of Turku, Finland}.\\
        Emails: 
        \{sier.s.ha, honghao.h.du, jian.j.song, xianjia.yu, tovewe\}@utu.fi\\[+6pt]
    }
}

\maketitle

\begin{abstract}
Modern lidar systems can produce not only dense point clouds but also 360 degrees low-resolution images. This advancement facilitates the application of deep learning (DL) techniques initially developed for conventional RGB cameras and simplifies fusion of point cloud data and images without complex processes like lidar–camera calibration. Compared to RGB images from traditional cameras, lidar-generated images show greater robustness under low-light and harsh conditions, such as foggy weather. However, these images typically have lower resolution and often appear overly dark. While various studies have explored DL-based computer vision tasks such as object detection, segmentation, and keypoint detection on lidar imagery, other potentially valuable techniques remain underexplored. This paper provides a comprehensive review and qualitative analysis of DL-based colorization and super-resolution methods applied to lidar imagery. Additionally, we assess the computational performance of these approaches, offering insights into their suitability for downstream robotic and autonomous system applications like odometry and 3D reconstruction.

\end{abstract}



\section{Introduction}\label{sec:introduction}
Lidar and camera are extensively used as primary sensors for robotic and autonomous systems~\cite{jiang2022autonomous, sier2023benchmark}. While deep learning (DL) techniques have matured significantly for processing RGB images from cameras, their application to lidar point clouds remains limited due to the high computational complexity and the relatively sparse semantic content of point clouds beyond geometric structure. 
Recent advancements in lidars enable the generation of images alongside point clouds by encoding the reflectivity from either ambient light or emitted laser signals~\cite{angus2018lidar}. These lidar-generated images are more robust to motion blur, illumination changes, and adverse weather conditions such as fog. Furthermore, they simplify the fusion of images and lidar point clouds, reducing the complexity of online or offline lidar-camera fusion. However, these lidar-generated images are typically low resolution and appear dark challenging their further processing. 

\begin{figure}[t]
    \centering
    \includegraphics[width=0.45\textwidth]{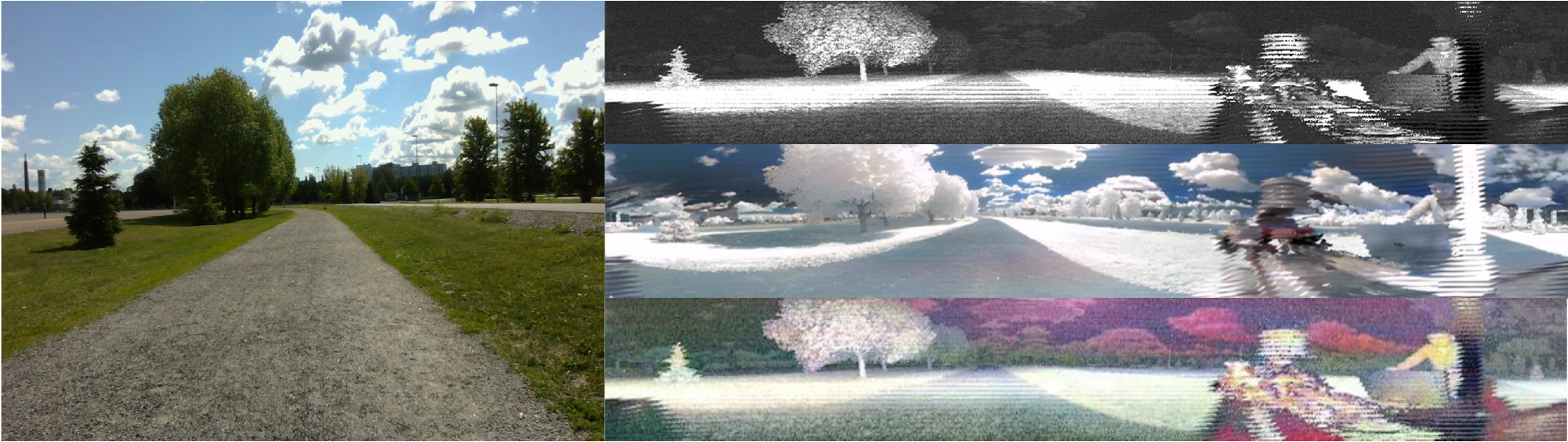}
     \caption{DL-based super-resolution and colorization results for lidar image: RGB (left), lidar signal, colorized near-IR, and colorized signal images (right, top to bottom)}
    \label{fig:result-sup-colorization}
\vspace{-1eM}
\end{figure}

Nowadays, studies have begun exploring the adaptation of DL-based techniques that were originally designed for conventional RGB imagery to lidar-generated images, including applications in object detection, segmentation, and keypoint extraction~\cite{yu2023general,zhang2023lidar}. Nevertheless, this area remains underexplored, and other potentially valuable DL approaches, such as super-resolution and colorization, have not been thoroughly evaluated. Super-resolution could enhance the spatial detail of panoramic lidar images, while colorization may enrich pixel-level information, offering potential benefits for tasks such as odometry and 3D reconstruction.
Fig.~\ref{fig:result-sup-colorization} illustrates our evaluation of one of the DL-based super-resolution and colorization approaches applied to lidar images.
The image on the left is an RGB image, while the images on the right (top to bottom) include one type of raw lidar image and its enhanced version via DL based super resolution and colorization.

To address these gaps, we present a comprehensive review of state-of-the-art DL-based super-resolution and colorization methods, surpassing the scope of prior surveys (Section~\ref{sec:dl-res-col}). Building on this foundation, we qualitatively evaluate several representative approaches on lidar-generated images from both indoor and outdoor datasets (Section~\ref{sec:METHODOLOGY}). We report the runtime performance of each method, providing insights for their potential deployment in real-world robotic systems.

\begin{table*}[t]
    \centering
    \caption{ Brief Comparison of Deep Learning-based Colorization Models } 
    \renewcommand{\arraystretch}{1.1}
    \resizebox{\linewidth}{!}{%
    \begin{tabular}{lccclc}
    \hline
    \textbf{Model} & \textbf{Indoor} & \textbf{Outdoor} & \textbf{Code} & \textbf{Description} & \textbf{Language compatible}\\ \hline
    \textit{DeOldify~\cite{antic2019deoldify}} & $\surd$ & $\surd$ & $\surd$ & Effectiveness in restoring and colorizing old black-and-white photos & Python3 \\
    \textit{DDColor~\cite{kang2023ddcolor}} & $\surd$ & $\surd$ & $\surd$ & Utilizes diffusion models for high-quality image colorization & Python3 \\
    \textit{BigColor~\cite{kim2022bigcolor}} & $\surd$ & $\surd$ & $\surd$ & Designed for large-scale, high-resolution image colorization & Python3 \\ 
    \textit{BigGAN~\cite{brock2018large}} & & $\surd$ & & Generating high-resolution, high-fidelity natural images \\
    \textit{PearlGAN~\cite{luo2022thermal}} &  & $\surd$ & $\surd$ & Focuses on colorizing thermal and infrared images using GANs, effective in outdoor environments & Python3\\
    \textit{I2V-GAN~\cite{I2V-GAN2021}} &  & $\surd$ & $\surd$ & GAN-based model for colorizing infrared to visible spectrum images & Python3\\
    \textit{Colorful Image Colorization~\cite{zhang2016colorful}} & $\surd$ & $\surd$ & $\surd$ & One of the first CNN-based colorization models, introducing probabilistic color assignment & Python3\\
    \textit{Let there be Color~\cite{IizukaSIGGRAPH2016}} & $\surd$ & $\surd$ & $\surd$ & A fully automatic image colorization method that works on both grayscale and natural images using deep learning & Lua\\
    \textit{DISCO~\cite{XiaHWW22}} & $\surd$ & $\surd$ & $\surd$ & Uses deep neural networks to provide automatic colorization with a focus on preserving detail and texture & Python3\\ 
    \textit{Deep Koalarization~\cite{DBLP:journals/corr/abs-1712-03400}} & $\surd$ & $\surd$ & $\surd$ & Combines a pre-trained VGG network with a colorization model for realistic colorization & Python3\\ 
    \textit{Palette~\cite{paltte}} & $\surd$ & $\surd$ & $\surd$ & A versatile colorization model that leverages palette-based techniques to achieve high-quality results across various image types & Python3\\
    \textit{ChromaGAN~\cite{vitoria2020chromagan}} & $\surd$ & $\surd$ & $\surd$ & GAN-based model that emphasizes perceptual loss for natural colorization results & Python3\\ 
    \textit{InstColorization~\cite{Su-CVPR-2020}} & $\surd$ & $\surd$ & $\surd$ & An instance-aware colorization model that handles complex scenes by focusing on individual object instance & Python3\\
    \textit{SCGAN~\cite{zhao2020scgan}} & $\surd$ & $\surd$ & $\surd$ & A self-consistent GAN-based colorization approach that ensures color consistency across different regions of an image & Python3\\ \hline
    \end{tabular}
    }
    \label{tab:colorization_models}   
    \vspace{-1em}
\end{table*}

\section{DL based super-resolution and colorization}\label{sec:dl-res-col}

\subsection{DL-based Colorization}\label{sec:dl-col}

DL-based colorization is commonly used in image restoration and thermal infrared image colorization. Table~\ref{tab:colorization_models} shows the existing approaches we have reviewed with the consideration of applied environments, implementation availability, programming language compatibility, and a brief description.



DeOldify~\cite{antic2019deoldify} is a DL-based tool that is used to colorize and restore old black and white images and videos using the NoGAN techniques.
The generator network learns to add colour to the greyscale image, while the discriminator network learns to differentiate between the real colour image and the generator-generated image. Compared to the traditional GAN model, this model prioritizes natural colors and performs well and consistently on landscapes. 

The core concept of the DDColor~\cite{kang2023ddcolor} model is the use of two distinct decoders for image colorization. The Pixel Decoder focuses on recovering spatial details and image structure, while the Colour Decoder learns semantically aware color representations from visual features at multiple scales. By combining the outputs of both decoders, DDColor aims to produce more natural and vivid colors, especially in complex scenes with multiple objects.

BigColor~\cite{kim2022bigcolor} serves as a precursor to the DDColor model, which builds upon and refines the foundations established by BigColor. This model introduces a novel approach based on BigGAN~\cite{brock2019large}, expanding the representation space and offering a range of coloring results.

BigGAN is a colorization model for generating high-resolution, high-fidelity natural images based on Large Scale GAN Training for High Fidelity Natural Image Synthesis~\cite{brock2019largescalegantraining} training. The framework consists of a generator and a discriminator. The generator creates images, the discriminator evaluates them against real images, and then iteratively trains to generate high quality images. 

PearlGAN~\cite{luo2022thermal} is tailored for converting nighttime thermal infrared (NTIR) images into daytime color (DC) images. It excels at producing high-quality colorization for NTIR images of open roads. PearlGAN is restricted to outdoor scenes, and its colorization capabilities are primarily focused on roads and trees, particularly in lidar-generated images.

I2V-GAN~\cite{I2V-GAN2021} is designed for unpaired infrared-to-visible video translation. Like PearlGAN, it is specialized for outdoor scenes and works with infrared images. I2V-GAN facilitates the conversion of infrared video data into visible spectrum equivalents, although it shares similar limitations in its applicability to specific scene types and environments.

Let There Be Color~\cite{IizukaSIGGRAPH2016} is a Convolutional Neural Network (CNN) trained using paired grayscale and color images. The model learns to colorize grayscale images by leveraging the correlations between the grayscale inputs and their corresponding color images. It employs a deep architecture to capture complex patterns and color distributions, resulting in visually appealing colorization of grayscale images. The approach effectively handles a variety of image types, providing robust colorization across different scenes. 

DISCO~\cite{XiaHWW22} provides a new method of colouring images by separating colour representation from spatial information, helping to produce images with vibrant and realistic colors. 

Deep Koalarization~\cite{DBLP:journals/corr/abs-1712-03400} combines CNNs with the advanced Inception-ResNet-v2 architecture, allowing it to capture complex details and nuances in images. The model leverages Inception modules for efficiently handling multi-scale features, while ResNet's residual connections enable deeper networks to be trained without encountering gradient vanishing issues. The model's complexity and high computational demands require significant processing power and memory. Paltte~\cite{paltte} is similar to Deep Koalarization, as it reproduces the full model.

ChromaGAN~\cite{vitoria2020chromagan} by utilizing an adversarial framework, ChromaGAN's network of generators and discriminators work in tandem to produce high-quality color images. Where the generator network learns to generate color images from grayscale inputs, the discriminator network evaluates the veracity of these colorizations. Like many GAN-based models, ChromaGAN requires significant computational resources and is a complex and time-consuming process.

Instance-aware Image Colorization~\cite{Su-CVPR-2020} Enhance coloring accuracy and fidelity by incorporating instance-level perception into the process. The model distinguishes between different instances of the same object class in an image, enabling more accurate and consistent application of color. The complexity of the model and instance-level segmentation increases the computing time.

SCGAN~\cite{zhao2020scgan} is the use of saliency maps to inform the generator network within the GAN framework to ensure that the generated colors are visually appropriate, which is particularly beneficial for images with complex compositions or where certain objects should stand out.




\begin{table*}[t]
    \centering
    \caption{Brief Comparison of Deep Learning based Super Resolution Models } 
    \renewcommand{\arraystretch}{1.2}
    \resizebox{\linewidth}{!}{%
    \begin{tabular}{lcllllc}
    \toprule
    \textbf{Model} & \textbf{Code} & \textbf{Architecture} & \textbf{Training Data} & \textbf{Strengths} & \textbf{Weaknesses} & \textbf{Language compatible} \\ 
    \midrule
    \textit{SRCNN~\cite{dong2015image}} &  & 3-layer CNN & ImageNet & Simple, Effective & Limited long-range dependency capture \\
    \textit{VDSR~\cite{kim2016accurate}} &  & 20-layer CNN & ImageNet & Deep architecture, Residual learning & High computational cost \\
    \textit{SRGAN~\cite{ledig2017photo}} &  & CNN with GAN & ImageNet, DIV2K & High perceptual quality & Potential artifacts \\
    \textit{DRRN~\cite{Tai-DRRN-2017}} &  & Recursive Residual Network & ImageNet, DIV2K & Parameter efficient, Deep architecture & High computational cost \\
    \textit{CARN~\cite{ahn2018fast}} & $\surd$ & Cascading Residual Network & DIV2K & Lightweight, Fast inference & Slightly lower accuracy & Python3\\
    \textit{SwinIR~\cite{liang2021swinir}} & $\surd$ & Transformer-based & DIV2K, Flicker2K & High performance, Handles large scale factors & High computational cost & Python3 \\  
    \textit{SCUNet~\cite{zhang2023practical}} & $\surd$ & CNN with Spatially Consistent Normalization & DIV2K & Maintains spatial consistency, Effective edge preservation & High computational resources, Training complexity & Python3 \\ 
    \textit{ESRGAN~\cite{wang2018esrgan}} & $\surd$ & GAN with Residual-in-Residual Dense Blocks & DIV2K, Flicker2K & Realistic textures, Superior perceptual quality & High computational cost, Can produce artifacts & Python3 \\ 
    \textit{DCSCN~\cite{yamanaka2020fastaccurateimagesuper}} & $\surd$ & CNN and Sparse Coding & CIFAR-10, ImageNet & Efficient high frequency detail capture & Struggles with complex images & Python3 \\ 
    \textit{CAT~\cite{chen2022cross}} & $\surd$ & Cross-Attention Transformer & DIV2K, Flickr2K & Strong cross-modal learning, High fidelity & Computationally expensive & Python3 \\
    \textit{Perceptual Losses~\cite{johnson2016perceptual}} &  & CNN with Perceptual Loss & MS-COCO, ImageNet & High-quality image generation, Real-time performance & Dependent on pre-trained networks, Generalization limits\\ 
    \textit{SinGAN~\cite{rottshaham2019singan}} & $\surd$ & GAN with Single Image Learning & Single Image & Effective for artistic enhancement, Versatile & Training on each image required, Limited scalability & Python3 \\ 
    \textit{DnCNN~\cite{zhang2017beyond}} & $\surd$ & Deep CNN with Batch Normalization & ImageNet, BSD400 & Effective denoising, Simple architecture & Limited to small scale factors & Python3 \\ 
    \textit{SR3~\cite{9887996}} & $\surd$ & Diffusion Model for SR & CelebA-HQ, FFHQ & Generates realistic and high-quality images & High computational cost, Slow inference & Python3 \\
    \bottomrule
    \end{tabular}
        }
    \label{tab:sr_models} 
    \vspace{-1em}
\end{table*}

\subsection{DL-based Super Resolution}\label{sec:dl-res}

DL has greatly advanced the development of image super-resolution, enabling the models to generate high-quality magnified images from low-resolution inputs. This part provides a comparative analysis of various DL super-resolution models, describing their architecture, performance,  programming language compatibility, advantages, and disadvantages (Table~\ref{tab:sr_models}).

SRCNN~\cite{dong2015image} is one of the pioneering DL models for image super-resolution. Its three-layer CNN is designed for feature extraction, non-linear mapping, and reconstruction. Trained on ImageNet with mean square error (MSE) as the loss function, SRCNN aims to minimize the difference between the output and the ground truth image. It achieves strong results on datasets like Set5, Set14, and BSD. Owing to its shallow architecture, SRCNN struggles to capture long-range dependencies in complex images, limiting its overall performance.

VDSR~\cite{kim2016accurate} model builds on the SRCNN framework by employing a much deeper CNN with 20 layers. This increased depth enables VDSR to learn more complex mappings from low- to high-resolution images, significantly improving performance. To mitigate the vanishing gradient problem, the model incorporates residual learning. VDSR consistently outperforms SRCNN on benchmark datasets, showcasing the benefits of its deeper architecture for super-resolution tasks. However, the added depth results in higher computational costs.

SRGAN~\cite{ledig2017photo} was the first to apply generative adversarial networks (GANs) to super-resolution tasks. SRGAN comprises two neural networks: a generator that creates realistic high-resolution (HR) images from low-resolution (LR) inputs, and a discriminator that distinguishes between generated images and real HR photos. 
Nevertheless, the adversarial training can introduce artifacts, and achieving a balance between the generator and discriminator remains challenging.

DRRN~\cite{Tai-DRRN-2017} leverages deep recursive learning with recursive residual blocks to effectively capture dependencies across different scales. DRRN employs a recursive structure with shared parameters, allowing it to increase network depth without significantly inflating the number of parameters. The recursive design complicates the training process and can be computationally intensive.

CARN~\cite{ahn2018fast} is a ResNet-based image super-resolution model designed for real-time applications. By using cascading connections, CARN reduces computational cost while retaining high performance, enabling it to merge features across layers and capture intricate image details more effectively. Trained on large-scale super-resolution datasets, CARN strikes a balance between speed and accuracy, though its efficiency-focused design can lead to slight compromises in image quality compared to more complex models.

SwinIR~\cite{liang2021swinir} leverages the Swin Transformer architecture for tasks like super-resolution. By dividing images into smaller patches and analyzing them at multiple scales, SwinIR captures both local details and long-range dependencies. This method enhances the model's understanding of the image's global structure, although it increases computational complexity.

SCUNet~\cite{zhang2023practical} proposes a method to enhance image quality by removing noise without prior knowledge of its properties. This is done by integrating SCUNet with a strategic data synthesis process. It outperforms traditional and state-of-the-art models in denoising tasks, effectively eliminating noise while preserving details and textures. However, its high computational demands and reliance on synthetic data are key concerns for practical deployment.

ESRGAN~\cite{wang2018esrgan} improves upon the ESRGAN model by introducing Residual Dense Blocks (RRDB) and Perceptual Loss Functions to enhance image realism. ESRGAN trains generators and discriminators simultaneously, resulting in superior image quality. ESRGAN is computationally demanding, requiring substantial processing power, memory, and numerous iterations for optimal results. Additionally, ESRGAN can produce artifacts, particularly in images with complex textures.

DCSCN~\cite{yamanaka2020fastaccurateimagesuper} leverages deep CNNs, skip connections, and Network-in-Network (NIN) architectures to improve super-resolution image quality while maintaining computational efficiency. This enables the generation of high-quality images with fine details and textures. Despite its efficient design, the deep architecture and NIN layers still require considerable processing power and memory. Moreover, if not properly normalized, the model’s strong feature extraction may cause overfitting, affecting performance on unseen data.



CAT~\cite{chen2022cross} uses a cross-attention mechanism to aggregate features from different image regions, enabling recovery of fine details and textures often lost in traditional methods. It also allows the model to efficiently handle various restoration tasks, including denoising, deblurring, and super-resolution. However, the complexity and computational demands of transformer architectures remain challenging and require substantial processing power.

Perceptual Losses~\cite{johnson2016perceptual} utilizes a perceptual loss function to enhance the visual quality of generated images, using high-level feature differences from a pre-trained CNN, rather than relying on traditional pixel-by-pixel comparisons. Despite its advantages, the model's reliance on pre-trained networks for perceptual loss calculations may limit its adaptability across different datasets, as performance may depend on the characteristics of the specific network used.


SinGAN~\cite{rottshaham2019singan} generates images by training a generative adversarial network (GAN) on a single image, allowing it to produce diverse outputs that preserve the original image’s texture, color, and structure. While its single-image capability is an advantage, SinGAN is computationally intensive, and its results can lack diversity compared to models trained on large datasets. Moreover, the quality of generated images depends heavily on the original image’s characteristics.

DnCNN~\cite{zhang2021plug} employs a deep CNN combined with NIN architecture and skip connections. These connections help integrate NIN layers, enhancing DnCNN's ability to capture fine details and represent high-frequency components, improving texture and edge preservation in super-resolution images. However, without adequate regularization, DnCNN's strong feature extraction capabilities can lead to overfitting, potentially affecting performance on unseen data.


SR3~\cite{9887996} enhances image resolution in multiple stages, with each iteration refining the output of the previous one. This progressively improves quality, producing finer details and more accurate textures in the final high-resolution image. However, the iterative process increases computational demands, which may limit its use in real-time applications or resource-constrained environments.


The comparison of these models include evolution from simple CNNs to complex transformer-based architectures. The summary in Table~\ref{tab:sr_models} provides a clear visual comparison.



\section{Evaluation on lidar imagery}\label{sec:METHODOLOGY}



\paragraph{Dataset for Evaluation}\label{subsec:my_dataset}
We conduct all evaluations using open-source multi-modal LiDAR datasets~\cite{sier2023benchmark, qingqing2022multi}, focusing on data from the Ouster LiDAR. Its specifications are listed in Table~\ref{tabs:ouster-table}. Ouster provides dense point clouds and multiple image types: range, signal, and ambient images, encoding depth, IR intensity, and ambient light, respectively. We primarily use signal images, which have shown strong performance in prior work~\cite{zhang2023lidar}.

\begin{table}[htb]
\centering
\caption{Specifications of Ouster OS0-128.}
\label{tabs:ouster-table}
\resizebox{0.48\textwidth}{!}{%
\begin{tabular}{lcccc}
\toprule
 &
  \textbf{IMU} &
  \textbf{Type} &
  \textbf{Channels} &
  \textbf{Image Resolution}  \\
   \midrule
\textbf{Ouster OS0-64} &
  ICM-20948 &
  spinning &
  128 &
  $ 1024 \times 128 $ \\
  \bottomrule
  \toprule
  \textbf{FoV} &
  \textbf{Angular Resolution} &
  \textbf{Range} &
  \textbf{Freq} &
  \textbf{Points}\\
  \midrule
  $360^\circ \times 90^\circ$ &
  $V:0.7^\circ, H:0.18^\circ$ &
  50\,m &
  10\,Hz &
  2,621,440 pts/s \\ \bottomrule
\end{tabular}%
}
\vspace{-1ex}
\end{table}

The evaluation data sequences include indoor and outdoor environments. The outdoor environment is from the normal road and a forest, denoted as~\textit{Open road} and~\textit{Forest}, respectively. The indoor data includes a hall in a building and two rooms, denoted as~\textit{Hall (large)}, ~\textit{Lab space (hard)}, and~\textit{Lab space (easy)}, respectively. Our own~\textit{Forest} dataset was collected within a forested area. 


    

\paragraph{Analysis of Colorization Approaches}
Section~\ref{sec:dl-col} presented a comprehensive survey on the DL-based colorization approaches. Here, we qualitatively analyze the performance of the approaches on lidar-generated images (signal images), utilizing publicly available code repositories on GitHub. 
The evaluation methods were selected based on their implementation availability and ease of deployment on a standard laptop.
More specifically, the approaches examined include~\textit{BigColor, Colorful Image Colorization, DDColor, DeOldify, DISCO, InstColorization, Let there be color, PearlGAN}, as detailed in Table~\ref{tab:colorization_models}.

\paragraph{Analysis of Super-Resolution Approaches}
The review of the DL-based super-resolution approaches is in Section~\ref{sec:dl-res}. 
Among these approaches, we select the methods based on the same principle mentioned above: the availability of implementation and the simplicity of deployment for a standard laptop. 
The approaches included in the part are~\textit{CARN, SwinIR, DCSCN, ESRGAN, SCUNET}, as detailed in Table~\ref{tab:sr_models}. We qualitatively evaluate the super-resolution performance of these approaches on signal images. 
\paragraph{Hardware Information}
The evaluation was conducted using a Razer Blade 15 laptop by Ubuntu 22.04.4 LTS equipped with an Intel Core i7-12800H-20 processor, 16\,GB of RAM with a frequency of 4800\,MHz, and a GeForce RTX 3070 Ti GPU with 8\,GB of memory.


\section{Experimental Results}
This section presents the qualitative evaluation results of  DL-based super-resolution and colorization models. Following this analysis, the most suitable models in terms of both processing speed and output quality based on our experimental findings are selected. These models are then applied to our point cloud sampling approach for further performance analyzes in terms of accuracy and the number of points extracted.

\begin{figure*}[t]
    \centering
    \begin{subfigure}[ht]{0.48\textwidth}
        \centering
        \includegraphics[width=0.95\textwidth]{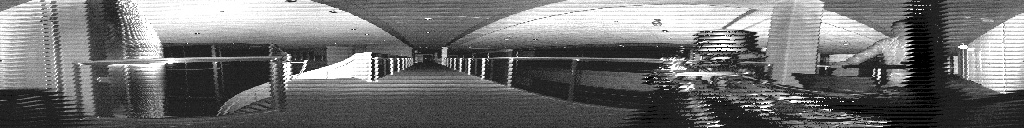}
        \caption{Raw image indoor}
        \label{fig:with-drift-raw-indoor} 
    \end{subfigure}
    \begin{subfigure}[ht]{0.48\textwidth}
        \centering
        \includegraphics[width=0.95\textwidth]{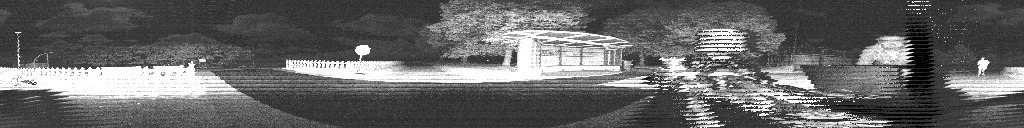}
        \caption{Raw image outdoor}
        \label{fig:with-drift-raw-outdoor} 
    \end{subfigure}
    \vspace{-1ex}
    \caption{Raw images of indoor and outdoor environment}
    \label{fig:drift-raw}
    \vspace{1ex}

    \begin{subfigure}[ht]{0.48\textwidth}
        \centering
        \includegraphics[width=0.95\textwidth]{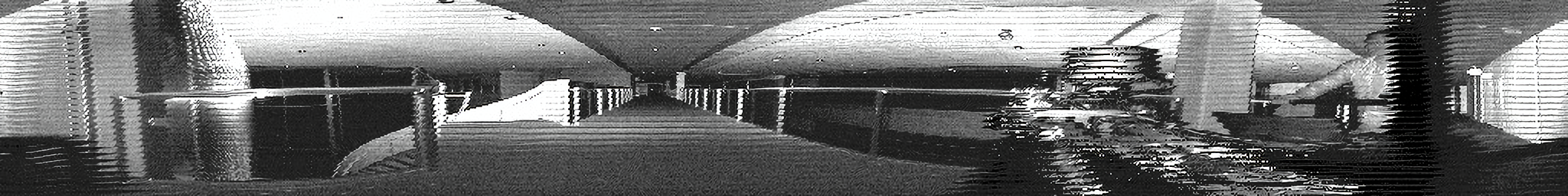}
        \caption{Example of CARN super-resolution indoor image}
        \label{fig:with-carn-sr-indoor} 
    \end{subfigure}
    \begin{subfigure}[ht]{0.45\textwidth}
        \centering
        \includegraphics[width=0.95\textwidth]{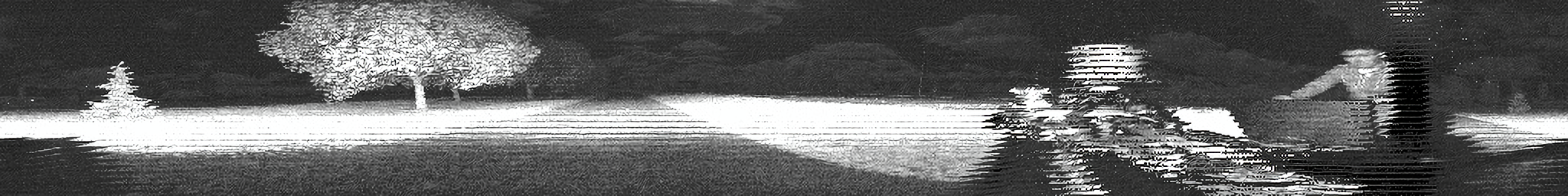}
        \caption{Example of CARN super-resolution outdoor image}
        \label{fig:with-drift-carn-sr-outdoor} 
    \end{subfigure}
%
    \begin{subfigure}[ht]{0.48\textwidth}
        \centering
        \includegraphics[width=0.95\textwidth]{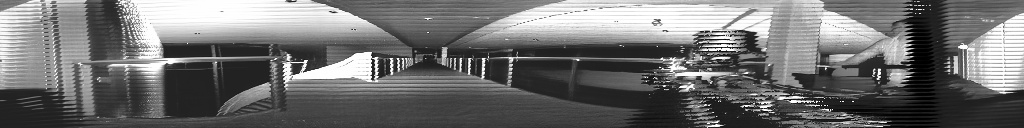}
        \caption{Example of SCUNET super-resolution indoor image}
        \label{fig:with-scunet-sr-indoor} 
    \end{subfigure}
    \begin{subfigure}[ht]{0.45\textwidth}
        \centering
        \includegraphics[width=0.95\textwidth]{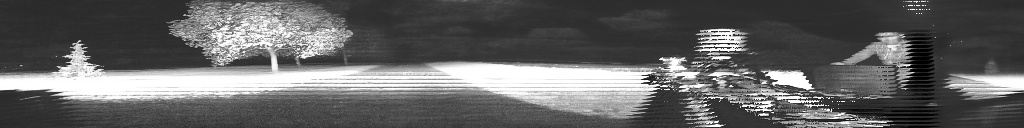}
        \caption{Example of SCUNET super-resolution outdoor image}
        \label{fig:with-drift-scunet-sr-outdoor} 
    \end{subfigure}
%
    \begin{subfigure}[ht]{0.48\textwidth}
        \centering
        \includegraphics[width=0.98\textwidth]{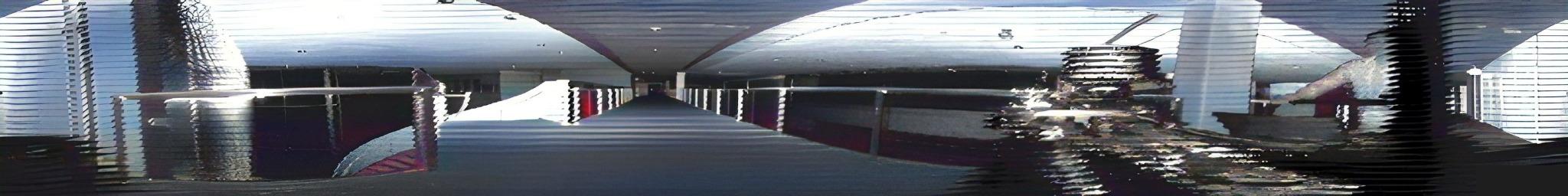}
        \caption{Example of DeOldify colorization indoor image}
        \label{fig:with-drift-deoldify-indoor} 
     \end{subfigure}
     \begin{subfigure}[ht]{0.45\textwidth}
         \centering
        \includegraphics[width=0.98\textwidth]{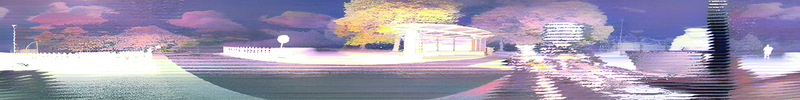}
        \caption{Example of DeOldify colorization outdoor image}
        \label{fig:with-drift-deoldify-outdoor} 
    \end{subfigure}
%
    \begin{subfigure}[ht]{0.48\textwidth}
        \centering
        \includegraphics[width=0.98\textwidth]{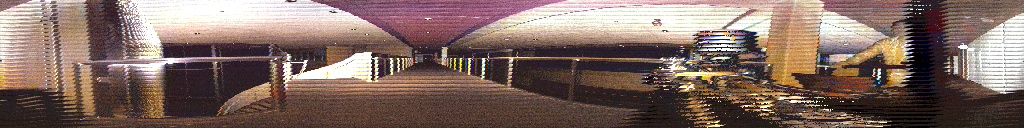}
        \caption{Example of DDColor colorization indoor image}
        \label{fig:with-drift-ddcolor-indoor} 
     \end{subfigure}
     \begin{subfigure}[ht]{0.45\textwidth}
         \centering
        \includegraphics[width=0.98\textwidth]{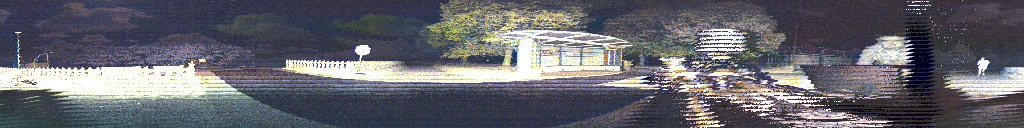}
        \caption{Example of DDColor colorization outdoor image}
        \label{fig:with-drift-ddcolor-outdoor} 
    \end{subfigure}
    \vspace{-1ex}
    \caption{Examples of using super-resolution and colorization on lidar-generated images}
    \label{fig:examples-sr_and_c}

    \vspace{-1em}

\end{figure*}

\paragraph{Super-Resolution of lidar imagery} 
The example processed images in Fig.~\ref{fig:with-carn-sr-indoor}\,-\,\ref{fig:with-drift-scunet-sr-outdoor}, like most super-resolution models, the final rendering for the signal image has a higher degree of sharpening or an overall smoother image than the raw image in Fig.~\ref{fig:drift-raw}. Table~\ref{tab:color_speed} shows the inference speeds of the multiple popular DL-based super-resolution models with input and output sizes illustrated. The DL-model~\textit{CARN} demonstrates relatively high result quality, as illustrated in Fig.~\ref{fig:with-carn-sr-indoor} and \ref{fig:with-drift-carn-sr-outdoor}, while also exhibiting fast processing speed.


\begin{table}[h]
    \caption{Comparison of running speed of each super-resolution model based on local environment}
    \label{tab:color_speed}
\resizebox{0.45\textwidth}{!}{%
    \begin{tabular}{@{}llcc@{}}
    \toprule
    \textbf{Model}  & \textbf{Input size}  & \textbf{Output size} & \textbf{Running speed (secs/image)}  \\ \midrule
    CARN   & (1024, 128) & (2048, 256) & 0.005 \\
    SwinIR & (1024, 128) & (4096, 512) & 2.217 \\ 
    DCSCN  & (1024, 128) & (2048, 256) & 0.238 \\ 
    ESRGAN & (1024, 128) & (4096, 512) & 2.667 \\
    SCUNET & (1024, 128) & (1024, 128) & 1.706 \\ \bottomrule
    \end{tabular}
}
\vspace{-1ex}
\end{table}

\paragraph{Colorization of LiDAR imagery}
Fig.~\ref{fig:with-drift-deoldify-indoor} and \ref{fig:with-drift-deoldify-outdoor} as well as Fig.~\ref{fig:with-drift-ddcolor-indoor} and \ref{fig:with-drift-ddcolor-outdoor} show the result of the two colorization models after coloring, which present different results due to the differences in the training datasets. Table~\ref{tab:color_speed} shows the inference speed of the multiple popular DL-based colorization models with input and output sizes illustrated.
The DL model
~\textit{DeOldify} produces good colorization results, as shown in Fig.~\ref{fig:result-sup-colorization} and Fig.~\ref{fig:with-drift-deoldify-indoor} and \ref{fig:with-drift-deoldify-outdoor}, while demonstrating relatively faster processing speed, as indicated in Table~\ref{tab:color_speed}.

\begin{table}[h]
    \caption{Comparison of running speed of each colorization model based on local environment}
    \label{tab:color_speed}
\resizebox{0.45\textwidth}{!}{%
    \begin{tabular}{@{}llcc@{}}
    \toprule
    \textbf{Model}           & \textbf{Input size}  & \textbf{Output size} & \textbf{Running speed (secs/image)} \\ 
    \midrule
    BigColor                    & (1024, 128) & (1024, 128) & 0.54 \\
    Colorful Image Colorization & (1024, 128) & (1024, 128) & 0.27 \\ 
    DDColor                     & (1024, 128) & (1024, 128) & 0.37 \\ 
    DeOldify                    & (1024, 128) & (1024, 128) & 0.23 \\
    DISCO                       & (1024, 128) & (1024, 128) & 3.27 \\
    InstColorization            & (1024, 128) & (256, 256)  & 0.05 \\
    PearlGAN                    & (1024, 128) & (1024, 128) & 0.28 \\ 
    \bottomrule
    \end{tabular}
}
\vspace{-1eM}
\end{table}






\section{Conclusion and Future Work}\label{sec:conclusion}
\begin{figure*}[t]
    \centering
    \begin{subfigure}[t]{0.48\textwidth}
        \centering
        \includegraphics[width=\textwidth]{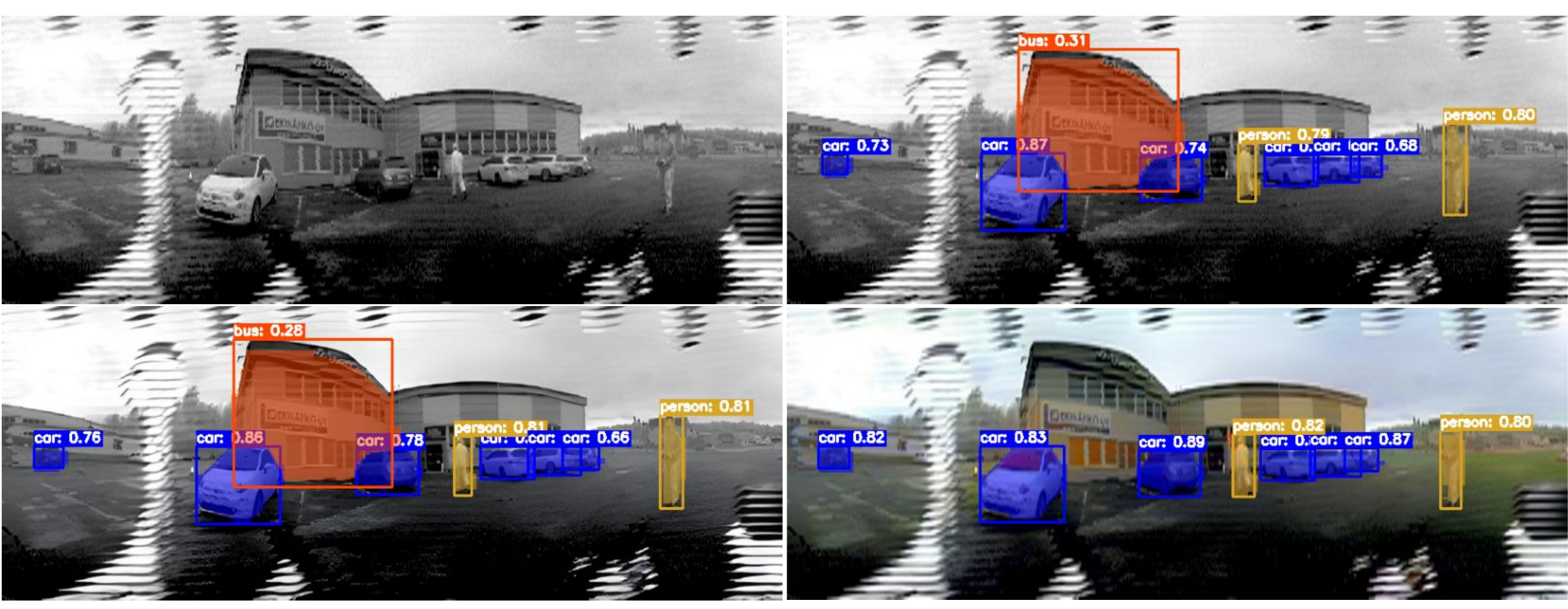}
        \caption{Instance segmentation on lidar near-infrared (NIR) images}
        \label{fig:seg_lidar_image_nir}
    \end{subfigure}
    \hfill
    \begin{subfigure}[t]{0.48\textwidth}
        \centering
        \includegraphics[width=\textwidth]{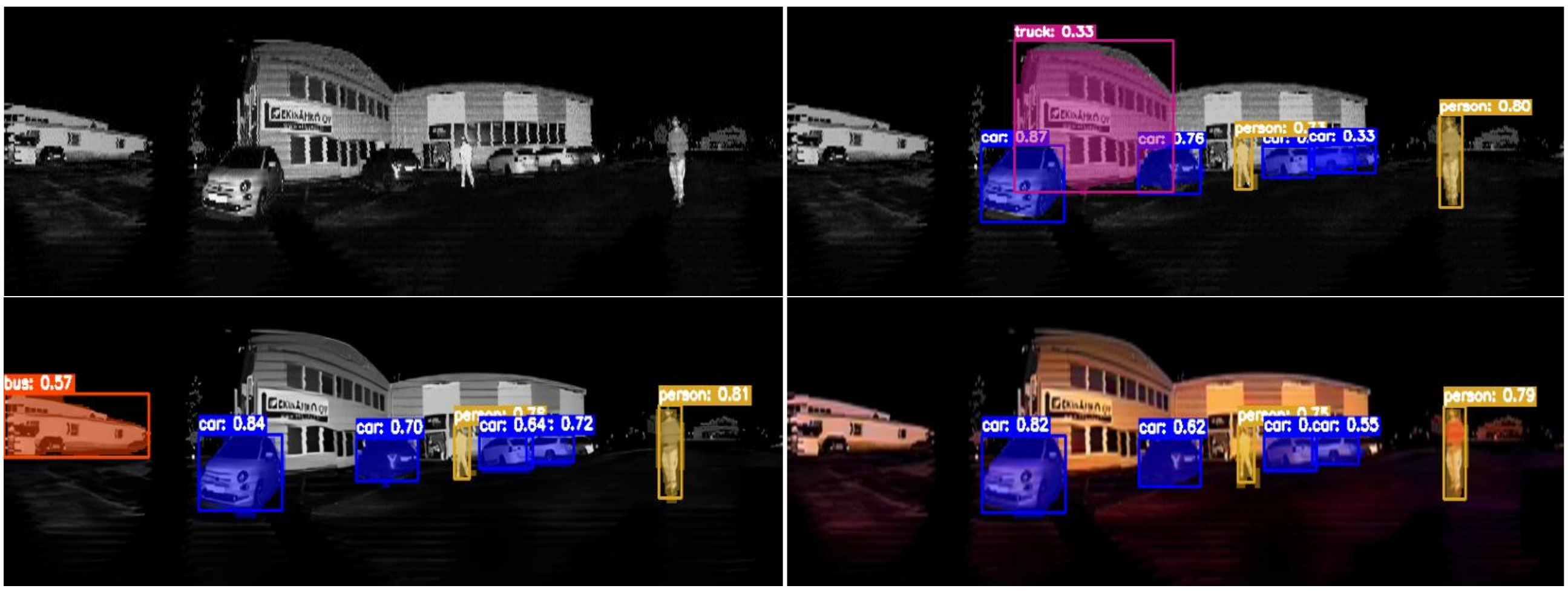}
        \caption{Instance segmentation on lidar reflectivity image}
        \label{fig:seg_lidar_image_reflect}
    \end{subfigure}
    \caption{Instance segmentation results for different lidar image types. Each image type (subfigure~\ref{fig:seg_lidar_image_nir} or ~\ref{fig:seg_lidar_image_reflect}) shows: original (top-left), result on original (top-right), after CRAN~\cite{ahn2018fast} super-resolution (bottom-left), and after super-resolution and Deoldify~\cite{antic2019deoldify} colorization (bottom-right).}
    \label{fig:seg_lidar_image_combined}
\end{figure*}

This paper provided a more comprehensive review of existing DL-based super-resolution and colorization methods than other literature. In this work, we have demonstrated the effectiveness of the DL-based super-resolution and colorization on lidar imagery. Additionally, we provide a runtime speed evaluation of those popular approaches on lidar imagery, posing the further application of these approaches in robotic and autonomous systems. 
%
Further research could focus on leveraging these methods to enhance lidar odometry or 3D reconstruction performance. Additionally, our experiments demonstrated that applying super-resolution and colorization techniques effectively reduced false detections made by the YOLOv11 instance segmentation model~\footnote{github.com/ultralytics/ultralytics}, as illustrated in Figure~\ref{fig:seg_lidar_image_combined}. For instance, the original NIR image in Figure~\ref{fig:seg_lidar_image_nir} (top-left) incorrectly identifies a building as a bus (top-right); however, this misclassification is successfully avoided in the enhanced image (bottom-right). A similar improvement is also evident in the lidar reflectivity image shown in Figure~\ref{fig:seg_lidar_image_reflect}. Although a systematic analysis was not conducted, these preliminary observations hold potential value for researchers working in related fields.





\section*{Acknowledgment}

This research is supported by the Research Council of Finland's Digital Waters (DIWA) flagship (Grant No. 359247) and  the DIWA Doctoral Training Pilot project funded by the Ministry of Education and Culture (Finland).

\bibliographystyle{abbrv-unsrt}
\bibliography{bibliography}

\end{document}